# Comprehensive performance evaluation of YOLOv12, YOLO11, YOLOv10, YOLOv9 and YOLOv8 on detecting and counting fruitlet in complex orchard environments

Ranjan Sapkota[a,*], Zhichao Meng[b], Martin Churuvija[c], Xiaoqiang Du[d], Zenghong Ma[d], Manoj Karkee[a,**]

[a] Biological and Environmental Engineering, Cornell University, Ithaca, 14850, NY, USA
[b] College of Optical and Mechanical Engineering, Zhejiang A&F University, Hangzhou, 311300, China
[c] Center for Precision & Automated Agricultural Systems, Washington State University, Prosser, 99350, WA, USA
[d] Zhejiang Sci-Tech University, School of Mechanical Engineering, Hangzhou, 310018, China

## HIGHLIGHTS

- We compare YOLOv8 - YOLOv12 for green fruitlet detection in complex, green-dominated orchard environments.
- An in-field counting efficiency analysis of the five latest YOLO models was performed under real-world orchard conditions.
- A cross-variety evaluation of detection and counting performance was conducted on four apple cultivars: SciFresh, SciLate, Honeycrisp, and Cosmic Crisp.
- Dual-sensor validation of model performance was achieved using both machine vision sensors and smartphone-based imaging systems.

## ARTICLE INFO

## ABSTRACT

This study systematically conducted an extensive real-world evaluation of all configurations of You Only Look Once (YOLO)-based object detection algorithms, including YOLOv8, YOLOv9, YOLOv10, YOLO11, and YOLOv12. Models were assessed using precision, recall, mean Average Precision at 50 % Intersection over Union (mAP@50), and computational efficiency across pre-processing, inference, and post-processing stages for detecting immature green fruitlets in commercial orchards. Field-level fruitlet counting was also validated using images captured with both Intel RealSense and iPhone 14 Pro Max sensors. YOLOv12l achieved the highest recall (0.900), while YOLOv10x and YOLOv9 GELAN-c reported the top precision scores of 0.908 and 0.903, respectively. YOLOv9 GELAN-base and GELAN-e achieved the highest mAP@50 (0.935), followed by YOLO11s (0.933) and YOLOv12l (0.931). In counting validation, YOLO11n demonstrated superior accuracy, with RMSE values of 4.51–4.96 and MAE values of 3.85–7.73 across four apple varieties. Sensor-specific training on Intel RealSense further improved detection performance. YOLO11n also recorded the fastest inference speed (2.4 ms), outperforming YOLOv8n, YOLOv9 GELAN-s, YOLOv10n, and YOLOv12n, affirming its suitability for real-time orchard applications.

## 1. Introduction

Object detection is a fundamental task in computer vision that enables automated systems to identify, classify, and locate objects within images or video streams [1,2]. This capability is critical for a wide range of applications, from autonomous driving and surveillance to robotics, healthcare, and agricultural automation [3]. By recognizing objects in real time, detection systems can support rapid decision-making and improve operational safety and efficiency in these applications [4]. In complex, real-world environments, object detection systems face various challenges such as lighting variations, scale changes, occlusions, and cluttered backgrounds. Overcoming these hurdles is essential for developing robust and reliable detection algorithms that work effectively under diverse conditions.

* Corresponding author.
** Corresponding author.
*E-mail addresses:* rs2672@cornell.edu (R. Sapkota), mk2684@cornell.edu (M. Karkee).

Object detection in commercial orchards is the foundation to develop agricultural automation and robotics solutions for labor-intensive tasks such as harvesting, thinning, and pruning [5,6]. One such labor-intensive operation in apple orchards is thinning green fruitlets in their early growth stage (called fruitlets), which is crucial due to its role in enhancing crop yield and quality. Automating the fruitlet thinning process is essential to minimize the dependence on rapidly depleting farm labor, which requires a robust machine vision system for fruitlet detection and localization in orchard environments [7].

Most of the tree fruit crops often set a greater number of fruit (as depicted in Fig. 1a) per tree than the desired number, which causes fruit-to-fruit competition for water, sunlight, and nutrients, resulting in inadequate exposure of the fruits to the sun, less space to grow, and overall reduced fruit quality [7]. Additionally, too many fruits can result in reduced cold hardiness, broken tree limbs, and depleted tree reserves [8]. Fruitlet thinning in the commercial production of tree fruit crops such as apples, kiwifruit, pears, peaches, and plums has been practiced for thousands of years to address these challenges and ensure optimal fruit size and quality [9,10].

Apples are the third most consumed fruit in the United States, with an average annual production of 4.6 million tons, making the U.S. the world's second-largest producer [11,12]. Approximately 382,000 acres are dedicated to apple cultivation, supporting exports of 42 million bushels and contributing an estimated $21 billion to the economy annually [12]. Labor shortages have become a critical challenge, as nearly 1.1 million crop-production workers face increasing demand, aging workforce demographics, and urban migration trends. Fruitlet thinning remains one of the most labor-intensive tasks in apple orchards, exacerbated by rising labor costs three times higher on specialty crop farms than the national average [13]. The COVID-19 pandemic further disrupted labor availability, resulting in an estimated $309 million loss in agricultural output from March 2020 to 2021 [14,15]. These challenges underscore the urgent need for robotic solutions to ensure sustainable and cost-effective apple production.

In commercial apple orchards, the demand for labor-intensive fruitlet thinning peaks in summer. Fig. 1a demonstrates the over-crowding of fruitlets on a tree branch, where each flower cluster has produced several fruitlets. Thinning is required to reduce this crowding, optimizing the number of fruitlets in each branch based on branch diameter, sunlight, space, and water. Fig. 1b shows workers manually thinning apple clusters, a process that is both time-consuming and laborious, requiring a significant number of seasonal workers.

As the U.S. agricultural workforce continues to decline, maintaining competitiveness in the global market necessitates the adoption of labor-saving technologies. Agricultural robots, which can replicate human tasks such as fruitlet thinning during the early growing season, offer a promising solution to labor-intensive operations [16]. In addition to reducing dependence on manual labor, automated and robotic systems for fruit thinning can also contribute to enhancing worker health and safety. Occupational health studies highlight the significant risks associated with manual labor in orchards. For instance, May et al. [17] reveal that farmworkers suffer from a high incidence of musculoskeletal injuries, skin diseases, and other health issues, with the agricultural sector experiencing a fatality rate seven times the national average. Further supporting this need, the study by Earle-Richardson et al. [18] documents the detrimental effects of prolonged physical labor on migrant orchard workers. The study by Kataoka et al. [19] demonstrates that such labor leads to significant decreases in muscle strength within a single workday.

The first crucial step in automating fruit thinning is developing a robust vision system capable of detecting green fruitlets during the early stage in commercial orchards. Green fruitlet detection is fundamentally an object detection challenge, a domain extensively studied within computer vision. Apple detection systems in orchard environments for potential automation have been reported for the last two decades [20]. In the past, many studies have reported apple detection models for supporting robotic harvesting using traditional image processing techniques such as image color segmentation in RGB, HSI (hue, saturation, and intensity), and/or HSL (hue, saturation, luminance) color spaces [21–23]. However, the accuracy of these traditional methods has been relatively low, and the approach is limited to specific environmental conditions such as uniform lighting and background conditions created artificially in orchards. It is challenging to implement these image processing algorithms to detect apples in a natural environment with high levels of occlusion, fluctuating illumination, and complex backgrounds [24].

During the last few years, machine learning (ML) has become widely adopted for object detection in agriculture, providing more robust detection capabilities with reduced levels of image pre-processing and feature extraction [25]. Some of the recent ML models used in detecting apples include Simple Linear Iterative Clustering (SLIC), Support Vector Machine (SVM) and Random Forest (RF) [26,27]. Additionally, deep learning (DL)-based models have become one of the most effective methods for performing object detection in agriculture as they provide end-to-end processing capability without any manual feature extraction. To perform apple detection in orchards, researchers have implemented DL models such as Faster Region-based Convolutional Neural Network (Faster R–CNN) models based on AlexNet and ResNet101 [24]. More recently, Kang et al. [28] have proposed a deep-learning-based detector called 'LedNet', which is able to perform real-time and accurate apple detection in orchards. The LedNet architecture utilizes the 3-level feature pyramid network (FPN) and Atrous Spatial Pyramid Pooling (ASPP)—a semantic segmentation module used in the feature processing block.

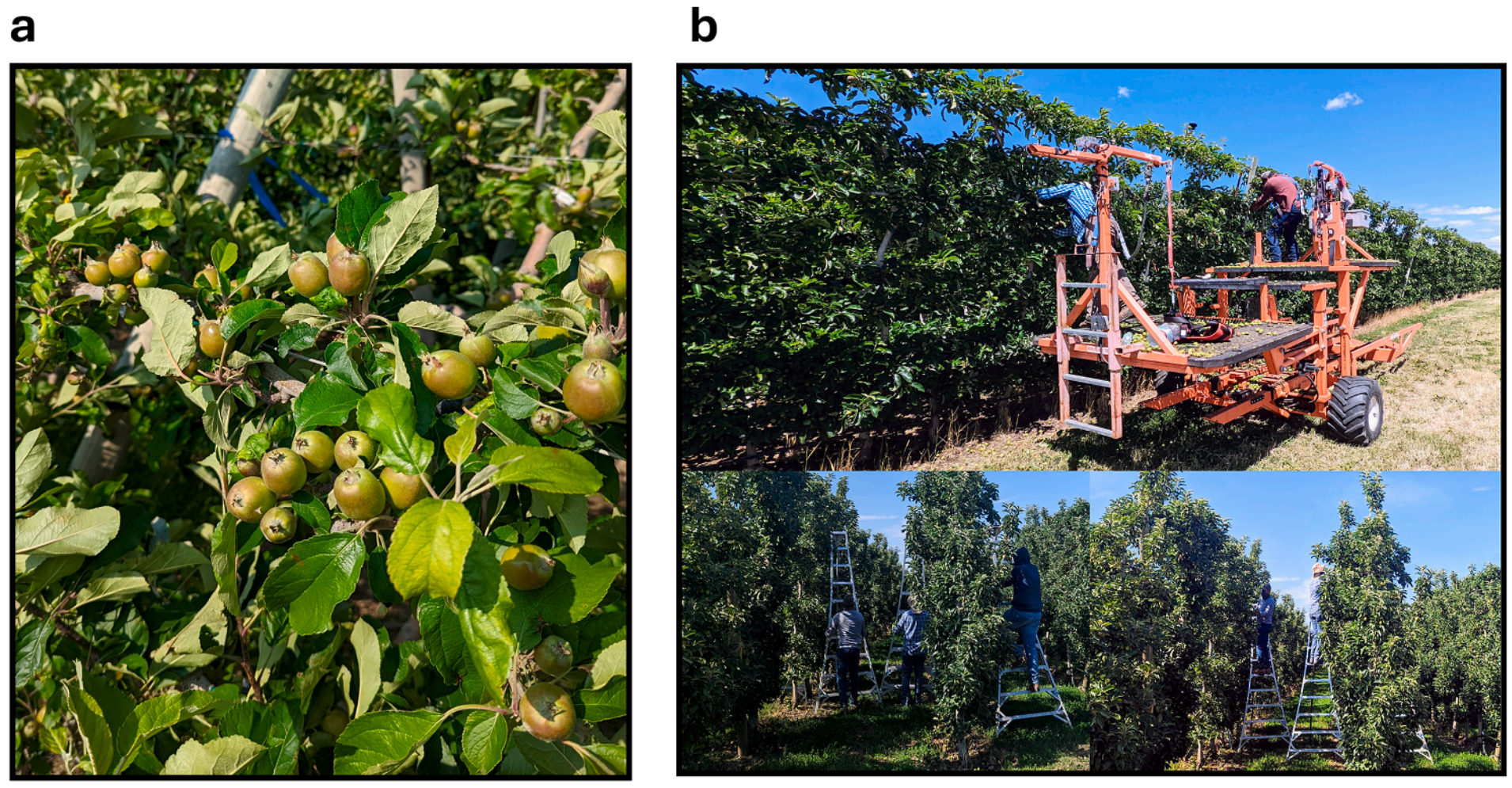

**Fig. 1.** Fruitlet thinning in commercial orchards. (**a**) High-density clusters of apple fruitlets on a SciLate apple tree during the peak thinning period in June 2022 (A commercial orchard in Prosser, WA), illustrating the typical over-cropping seen in commercial orchards. (**b**) Top: Laborers utilizing a height-adjustable platform for efficiently thinning fruitlets in various parts of the tree canopies. Bottom: A worker in a SciFresh apple orchard manually thinning excess fruitlets using an aluminum ladder, a common practice that highlights the labor-intensive nature of this crucial agricultural task.

Although researchers have been successful in detecting mature apples for robotic harvesting, there are only a few reported studies on the detection of apples during their early growing season for robotic thinning applications. During the early growing season in natural environments, both the color and light reflection of apples are similar to those of leaves, which poses a great challenge for accurately detecting apple fruitlets. Additionally, factors such as occlusion by leaves and branches, fruit overlap, and varying illumination make it difficult to detect fruitlets effectively [29,30].

Recently, to detect young apples, Tian et al. [31] implemented a YOLOv3 model called 'DenseNet' and reported an F1 score of 83.2 %. Additionally, to detect apples, Huang et al. [32] recently used an improved YOLOv3 model based on the CSPDarknet53 DL network. The study reported an inference speed of 8.6 ms in detecting immature apples. However, F1 and mAP achieved by the proposed model were only 0.65 and 0.67, respectively. Most recently, Wang et al. [33] presented a YOLOv5s-based deep learning technique to detect apple fruitlets before thinning in a natural environment. The authors of this study implemented transfer learning using a channel-pruned YOLOv5s object detection model and fine-tuned it to achieve a higher detection rate. However, the model deployed in this study was very small (1.4 MB), and the authors reported a false positive rate of 4.2 %.

Recent advancements in deep learning models for object detection have significantly enhanced fruitlet detection performance in field environments. The YOLO framework, modified for specific agricultural needs, has shown promising results. For instance, Jia et al. [34] reported that a tailored YOLO model improved the detection of green citrus in complex orchard environments achieving a substantially higher accuracy and speed compared to YOLOv4 [35]. Similarly, a lightweight version of YOLOv8 was developed for detecting pomegranate fruitlets, achieving a high precision with reduced model complexity and demonstrating suitability for mobile devices [36]. Additionally, a channel-pruned YOLOv5s model effectively detected apple fruitlets under varying conditions, outperforming several existing methods while maintaining a compact size suitable for potential use in mobile fruit thinning terminals [37].

Notably, an adaptation of the YOLOX-m model, optimized for robust feature extraction and enhanced by a feature pyramid network (FPN) integrated with an Atrous Spatial Pyramid Pooling (ASPP) module, has shown improvements in detecting fruitlets. The ASPP module-based network achieved an average precision of 64.3 % for apples and 74.7 % for persimmons, outperforming several common detection models and providing a solid reference for diverse fruit and vegetable detection tasks [34]. Similarly, the YOLO-P model, a modified version of YOLOv5 designed specifically for pear detection in orchards, incorporated advanced architectural changes such as ShuffleNet modules and a convolutional block attention module (CBAM). These modifications boost the feature extraction capabilities, allowing the model to perform well under challenging environmental conditions including complex backgrounds and varied lighting. This model not only supports rapid and precise pear detection but also offers insights for enhancing fruit detection systems in similar unstructured environments [38].

Moreover, a YOLOv4-based method specifically tailored for fig fruit detection in dense foliage has shown improvements over previous models including Faster R–CNN, emphasizing YOLO's capability in challenging conditions [39]. YOLOv8 has also been successfully applied to measure the size of immature green fruitlets using geometric shape fitting techniques, showcasing high precision and robustness against occlusions in orchard environments [7]. Furthermore, an enhanced YOLOv8 model has proven to be effective for detecting Yunnan Xiaomila peppers, integrating attention mechanisms and deformable convolutions to handle small targets against complex backgrounds [40]. Additionally, modifications to YOLOv5 have improved plum detection, adapting the algorithm to handle fruit occlusions and environmental variability [41]. A YOLOv7-based model equipped with attention mechanisms and specialized pooling has been shown to achieve high precision in detecting plums in natural settings [42].

In orchard automation, YOLO object detection models have been pivotal in enhancing the accuracy and efficiency of fruit detection [7,43,44], flower identification [45,46] and automated harvesting processes [6,39,47]. These models identify and classify fruits at various stages of ripeness, detect flowers with high precision, and enable efficient harvesting. The development of YOLO models has introduced significant improvements specifically tailored to meet the challenges faced in agricultural environments. For instance, the introduction of multi-scale predictions in YOLOv5 has improved the detection of small and clustered objects like flowers and young fruits, which has been crucial during the early stages of crop yield management [48].

The rapid advancement of YOLO-based models has significantly enhanced their precision and processing speeds, which are vital for implementing robotic solutions such as fruitlet thinning in orchards. Given these swift innovations, the continuous evaluation of newer models is essential to effectively translate these improvements into practical agricultural automation.

### 1.1. Objectives

This study provides a comprehensive evaluation of the latest YOLO versions: YOLOv12, YOLO11, YOLOv10, YOLOv9, and YOLOv8, for fruitlet detection in commercial apple orchards. We examined 26 configurations across these models utilizing a dataset of RGB images captured with an iPhone 14 Pro Max across four apple varieties: SciFresh, SciLate, Honeycrisp, and Cosmic Crisp. Each image was annotated with a total count of fruitlets, which was determined by combining the visible fruitlets in the image with the count of occluded fruitlets that were directly observed and recorded in the field at the time of image capture, providing a comprehensive ground-truth dataset for validation. The study further validates the top-performing models on an additional dataset captured with a dedicated machine vision sensor (Intel RealSense).

The specific contributions of this study are fourfold. First, we present a comprehensive evaluation of the most recent YOLO object detection models, tested across 26 distinct configurations to optimize performance in detecting green fruitlets in commercial apple orchards. These include YOLOv8 (five configurations), YOLOv9 (six configurations), YOLOv10 (six configurations), YOLO11 (five configurations), and YOLOv12 (four configurations). Second, we provide an in-depth analysis of performance through detailed examination of detection precision metrics, computational efficiency, and processing speeds measured at three stages of the deep learning workflow: pre-processing, inference, and post-processing. Third, we validated the models in-field for counting accuracy across four apple varieties excluded from the training set, thereby testing the generalizability and robustness of the approach under diverse agricultural conditions. Finally, the study integrates advanced smartphone technology by utilizing high-resolution RGB images captured with an iPhone 14 Pro Max, demonstrating the adaptability of cutting-edge smartphone imaging for real-world orchard applications. Collectively, these contributions highlight a rigorous evaluation pipeline, robust validation strategy, and practical deployment pathway toward precision fruit detection in orchard environments.

### 1.2. Evolution of object detection and the emergence of YOLO

The evolution of object detection with YOLO algorithms (as depicted in Fig. 2) has been driven by the quest for greater accuracy, speed, and robustness in real-world scenarios. Early methods relied on handcrafted features and classical machine learning techniques. Techniques such as sliding window methods [49] and the Viola–Jones detector [50] used features like Haar-like descriptors, histograms of oriented gradients, and local binary patterns to capture essential visual information. Although these methods laid the foundation for automated detection, they were limited by the reliance on manual feature engineering and exhaustive search processes, which made them computationally intensive and less adaptable to the intricacies of natural scenes. The advent of deep learning, particularly the introduction of convolutional neural networks (CNNs), revolutionized the field by enabling automated, hierarchical feature extraction [51]. Models such as R–CNN, Fast R–CNN, and Faster R–CNN significantly improved detection accuracy by learning complex representations directly from data. Simultaneously, the Single Shot MultiBox Detector (SSD) streamlined the process by integrating region proposal and detection in a single network pass, setting the stage for real-time performance [52]. Among these transformative developments, the YOLO series emerged as a breakthrough approach by unifying detection and classification into one efficient, end-to-end framework, dramatically reducing computation time while maintaining competitive accuracy [53].

The initial iterations of the YOLO series marked a departure from traditional multi-stage detectors. YOLOv1 introduced a grid-based approach that enabled the model to simultaneously predict bounding boxes and class probabilities, thereby establishing a new paradigm in real-time detection [53]. Building on this success, YOLOv2 improved image resolution and expanded the range of detectable classes, enhancing both accuracy and versatility. YOLOv3 further refined the approach by incorporating multi-scale predictions and a deeper network architecture, which enabled more effective detection of small objects and complex features [54]. As the series progressed, YOLOv4 and YOLOv5 introduced architectural innovations such as Cross-Stage Partial (CSP) networks and advanced data augmentation techniques that enhanced feature representation and inference speed [55,56]. Subsequent versions like YOLOv6 and YOLOv7 continued to optimize the balance between accuracy and computational efficiency through innovations like dynamic label assignment and refined architectural blocks [57,58]. Together, these early YOLO models laid a robust foundation for real-time object detection by addressing many limitations of earlier methods and setting a high standard for speed and precision.

### 1.3. Recent YOLO iterations (YOLOv8 to YOLOv12)

YOLOv8 builds on the advances of its predecessors by adopting a more efficient and anchor-free design. Its backbone leverages a refined CSP-based module to extract hierarchical features effectively, while its neck integrates a streamlined FPN to enhance multi-scale feature representation. The decoupled head separates the objectness, classification, and regression tasks, improving precision without incurring significant computational overhead. This design simplifies the detection pipeline and ensures adaptability to varying object sizes and aspect ratios, achieving a strong balance between detection accuracy and computational efficiency for real-time applications [3].

Likewise, YOLOv9 introduces significant improvements through the incorporation of programmable gradient information (PGI) and the Generalized Efficient Layer Aggregation Network (GELAN). The PGI mechanism ensures that robust and reliable gradients are maintained across the network, which aids in efficient weight updates and better convergence [59]. GELAN further enhances the model by providing flexible and efficient multi-scale feature aggregation, which is essential for accurately detecting objects in complex scenes. This architectural refinement allows YOLOv9 to address the challenges of information loss in deep networks and maintain high detection accuracy, particularly for smaller objects.

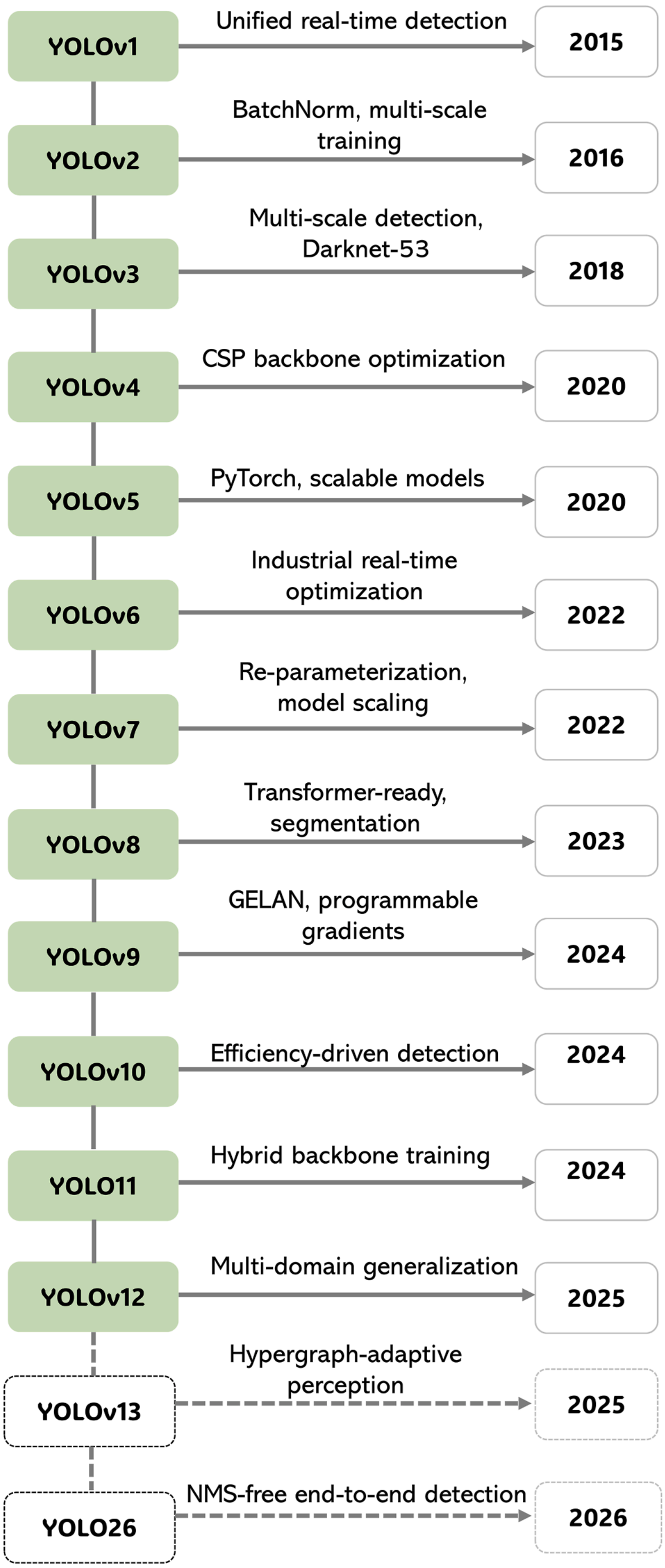

**Fig. 2.** Timeline diagram summarizing the evolution of YOLO object detection models from YOLOv1 (2015) to YOLOv12 (2025), highlighting key architectural and training innovations, and extending to post-YOLOv12 models (YOLOv13 and YOLO26) that represent emerging research directions toward end-to-end, NMS-free, and adaptive detection frameworks.

Building upon the innovations of YOLOv9, YOLOv10 further streamlines the object detection pipeline by eliminating the need for non-maximum suppression through a novel dual assignment strategy [60]. This approach reduces redundancy by directly assigning labels during training via one-to-many and one-to-one strategies, significantly cutting down on inference time. YOLOv10 also introduces lightweight classification heads and incorporates spatial-channel decoupled down-sampling, which minimizes information loss during feature extraction. Additionally, the rank-guided block design optimizes parameter utilization, ensuring that the network remains efficient even under the demands for real-time detection tasks. Furthermore, YOLO11 refines the object detection process through enhanced feature extraction and more robust detection capabilities. It introduces the C3k2 block to replace the C2f module, which significantly improves gradient flow and computational efficiency [3]. YOLO11 also integrates the Spatial Pyramid Pooling-Fast (SPPF) module to better capture multi-scale contextual information and employs the Convolutional block with Parallel Spatial Attention (C2PSA) to refine spatial feature representation. These enhancements allow YOLO11 to perform more accurate detection, particularly in scenarios involving occlusion and complex backgrounds [61].

Likewise, representing the state-of-the-art in the YOLO series, YOLOv12 adopts an attention-centric design that pushes the boundaries of real-time object detection [62]. Central to its innovation is the introduction of the Area Attention ($A^2$) module, which dynamically adjusts the receptive field to capture both global and local contextual cues with minimal computational expense. In parallel, the Residual Efficient Layer Aggregation Network (R-ELAN) refines feature aggregation by incorporating residual connections that ensure stable gradient flow and efficient information fusion. YOLOv12 also leverages additional optimizations, including Flash Attention and adaptive MLP ratios, to further enhance inference speed while maintaining high detection accuracy.

## 2. Methods

This study was conducted across commercial apple orchards in Prosser and Naches, Washington State, USA. It involved evaluating datasets of apple fruitlets before thinning from four varieties: SciFresh, SciLate, Cosmic Crisp, and Honeycrisp, as shown in Fig. 3. RGB images were acquired using a machine vision sensor, the Intel RealSense 435i (Intel Corporation, California, USA). These images were then manually labeled to prepare consistent datasets for training all configurations of YOLOv8, YOLOv9, YOLOv10, YOLO11, and YOLOv12 models. A total of 26 model configurations were examined (Fig. 3): five for YOLOv8, six for YOLOv9, six for YOLOv10, five for YOLO11, and four for YOLOv12. Each model was trained under standardized hyperparameter settings on the same computational system to ensure consistency and comparability. The trained models' performance was evaluated using the prepared datasets, followed by in-field counting validation using additional RGB images captured with an iPhone 14 Pro Max smartphone (Apple Inc., California, USA). This validation process aimed to assess the fruit counting accuracy of each model against manual ground-truth data. The validation set specifically included scenarios with occluded fruitlets. Furthermore, this validation step was designed to rigorously test the top-performing models in terms of speed and accuracy using images from different sensors.

### 2.1. Study site and data acquisition

The image dataset for this study was collected at Allan Brothers Orchard in Prosser, Washington State, USA, focusing on datasets of immature green fruitlets for training models. RGB images were acquired using an Intel RealSense D435i camera in June 2024. For validation, additional datasets were collected from different locations and times: including Cosmic Crisp apple images from the ROZA experiment station at Washington State University's Irrigated Agriculture Research and Extension Center (WSU IAREC) in Prosser, and Honeycrisp apple images from an orchard in Naches owned by Allan Brothers Fruit Company. Furthermore, all images were captured prior to the fruitlet thinning process conducted by orchard workers.

The details of each machine vision sensor used in data collection of this study are: (1) Intel RealSense D435i: This camera features a 2-megapixel RGB sensor capable of capturing high-quality images. Operating at a resolution of 1280 × 720 pixels, it provides a comprehensive view with a 69.4° horizontal and 42.5° vertical field-of-view, ensuring broad coverage in various environments. The Intel RealSense D435i is designed to be compact and lightweight, making it highly effective for capturing RGB data in diverse settings. (2) iPhone14 Pro Max: The iPhone 14 Pro Max features an advanced 48-megapixel RGB camera, which includes a 24 mm lens with an f/1.78 aperture and second-generation sensor-shift optical image stabilization, ensuring high-resolution image capture with enhanced clarity. The camera's field of view spans 120° horizontally and 90° vertically, supported by a seven-element lens to reduce optical aberrations. It offers up to 15x digital zoom and 4K video recording capabilities. Additionally, the integration of Photonic Engine and Smart HDR 4 technologies optimize performance in low-light conditions and dynamic range, making it suitable for detailed visual data collection in varying lighting environments.

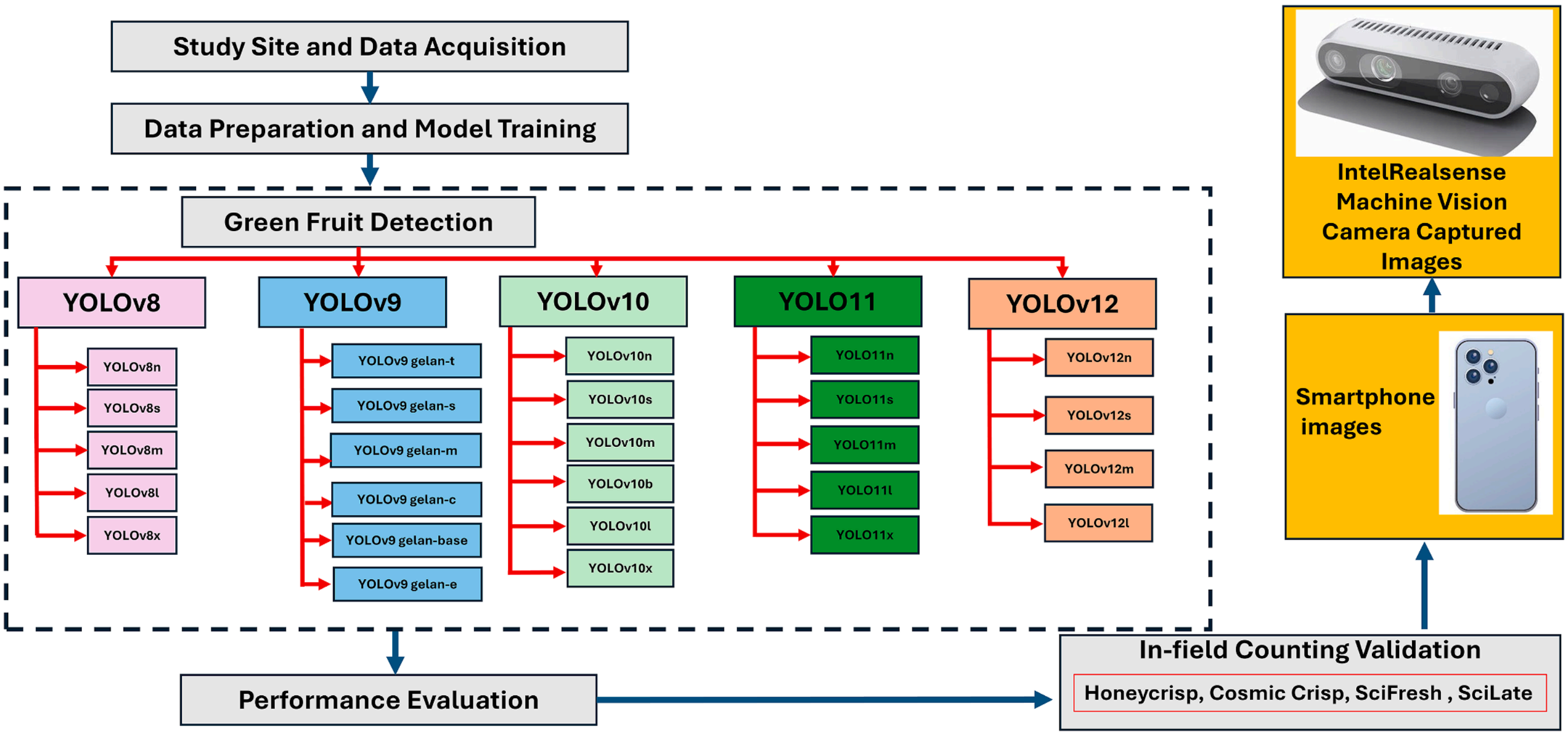

**Fig. 3.** Flow diagram of the proposed study comparing YOLOv8, YOLOv9, YOLOv10, YOLO11, and YOLOv12 models. The methodology includes data collection, model training, and validation across images collected with multiple sensors and from multiple apple varieties in commercial orchards.

To ensure a robust and diverse dataset for training object detection models, we collected a total of 1147 RGB images of immature apple fruitlets from the two commercial orchards under a wide range of environmental and operational conditions. Specifically, 592 images were acquired from a SciFresh apple orchard, and 555 images were collected from a SciLate apple orchard, both located in Prosser, Washington State. The collection aimed to capture a full spectrum of natural lighting conditions, including low-angle morning light, intense midday sunlight, diffuse lighting during overcast skies, and mixed-shadow conditions in partly cloudy weather. This range included variations in solar intensity, ambient shadows, and illumination contrast, simulating realistic field conditions encountered in commercial orchards. These intervals were chosen to represent a full spectrum of natural lighting conditions including strong shadows, high solar irradiance, and diffuse evening illumination. Data were acquired on clear sunny days (ambient light 75,000–90,000 lux), cloudy overcast days (10,000–20,000 lux), and partly cloudy conditions with intermittent light fluctuations. The camera was positioned at multiple horizontal and vertical angles (ranging from 30° to 90° zenithal tilt and 0°–180° azimuthal rotation) to mimic real-world robot-mounted viewpoints and simulate occlusions from leaves and branches. These measures ensured that the dataset encompassed a realistic distribution of variations in image brightness, background clutter, occlusion, and fruit orientation, thus supporting strong model generalizability under field conditions.

### *2.2. Data preparation and model training*

The images were manually annotated with bounding boxes using the online labeling platform provided by Roboflow. These images were allocated into training and validation at the ratio of 8:2, facilitated by Roboflow's distribution tools. No image pre-processing steps were undertaken in this study, as the objective was not to enhance any specific model but rather to compare and evaluate the performance of these models using raw data collected in natural orchard settings.

The computational analyses for this study were conducted on a high-performance workstation equipped with an Intel Xeon (R) W-2155 CPU, featuring a base clock speed of 3.30 GHz across 20 cores. This setup provided substantial processing power necessary for handling intensive data processing tasks. The workstation was also outfitted with NVIDIA Corporation GP102 [TITAN Xp] graphics cards, enhancing its ability to perform complex image processing and machine learning tasks efficiently. The system included a considerable storage capacity of 7.0 TB, facilitating extensive data management and analysis. It operated under Ubuntu 20.04.6 LTS, a robust and stable 64-bit operating system, using GNOME version 3.36.8 for its graphical interface and X11 as its windowing system.

The analysis of YOLOv8, YOLOv9, YOLOv10, YOLO11, and YOLOv12 encompassed a total of 26 configurations (five for YOLOv8, six for YOLOv9, six for YOLOv10, five for YOLO11, and four for YOLOv12). Each model configuration was thoroughly tested for precision, recall, mAP@50, and image processing speed. For the training of all models, a consistent configuration was rigorously maintained to ensure uniformity and comparability across experiments. Each model was trained for 700 epochs to allow sufficient time for learning and adaptation to the dataset's complexities. The batch size was set at 8, optimizing the balance between memory usage and processing speed. Images were resized to a uniform resolution of 640 × 640 pixels to standardize the input data size across all models. The Stochastic Gradient Descent (SGD) optimizer was employed to update model weights effectively, selected for its efficiency in handling large-scale and complex data. Additionally, the configuration threshold for confidence was set at 0.25, and the Intersection over Union (IoU) threshold was maintained at 0.7, criteria chosen to optimize the balance between precision and recall during object detection tasks.

The hyperparameter settings outlined in Table 1 provide a detailed framework for optimizing the training of YOLO models in the study. Key parameters such as the initial and final learning rates were set at 0.01, facilitating a controlled adjustment of learning throughout the training process. Momentum was maintained at 0.937 to ensure consistent updates across epochs, while a minimal weight decay of 0.0005 helped prevent overfitting. The training included a warmup phase of 3 epochs to stabilize the learning parameters early in the training. Loss adjustments were specifically tuned, with box loss, class loss, and objectness loss set at 7.5, 0.5, and 1.5 respectively, to balance the contributions of different components of the loss function. Image augmentation techniques such as hue, saturation, and value adjustments were precisely defined to enhance model robustness under varied lighting conditions common in orchard environments. Specific settings for geometric transformations like rotation, translation, and scaling were employed to simulate different orientations and sizes of objects, crucial for improving the model's ability to generalize across diverse scenarios. Table 1 also highlights the use of flipping and mosaic data augmentation to further enrich the training dataset, ensuring comprehensive exposure to potential real-world variations.

### *2.3. Performance evaluation*

To evaluate the detection capabilities of each configuration of YOLOv8, YOLOv9, YOLOv10, YOLO11, and YOLOv12, the metrics of precision, recall, and mean Average Precision at Intersection over Union (IoU) threshold of 0.50 (mAP@50) were employed. Precision was calculated as the ratio of true positive detections to the total predicted positives, given by equation (1), where TP denotes true positives and FP denotes false positives. Recall measured the ratio of true positive detections to the actual positives, formulated as equation (2), where FN represents false negatives. The mAP@50 was determined by averaging the precision across all recall levels for an IoU > 0.50. Additionally, the image processing speed for each model was analyzed and compared across three categories: pre-processing, inference, and post-processing. These evaluations were systematically conducted across 26 configurations of the YOLO models: YOLOv8m, YOLOv8s, YOLOv8l, YOLOv8x, YOLOv8c for YOLOv8; YOLOv9 GELAN-e, YOLOv9 GELAN-c, YOLOv9

**Table 1**
Hyperparameter Settings for YOLOv8, YOLOv9, YOLOv10, YOLO11 and YOLOv12 used in training YOLO models for fruitlet detection in commercial apple orchards.

| Hyperparameter | Value | Description |
|---|---|---|
| Initial Learning Rate (lr0) | 0.01 | Sets the starting learning rate. |
| Final Learning Rate (lrf) | 0.01 | Determines the learning rate at the end of training. |
| Momentum | 0.937 | Controls the momentum for the SGD optimizer. |
| Weight Decay | 0.0005 | Helps in regularizing and preventing overfitting. |
| Warmup Epochs | 3.0 | Number of initial epochs for learning rate stabilization. |
| Box Loss Gain (box) | 7.5 | Weight of the bounding box loss component. |
| Class Loss Gain (cls) | 0.5 | Weight of the class prediction loss component. |
| Definition Loss Gain (dfl) | 1.5 | Weight of the definition loss component. |

GELAN-s, YOLOv9 GELAN-t, YOLOv9 GELAN-m, and YOLOv9 GELAN for YOLOv9, YOLOv10m, YOLOv10s, YOLOv10l, YOLOv10x, YOLOv10c, YOLOv10d for YOLOv10, and YOLO11n, YOLO11s, YOLO11m, YOLO11l, and YOLO11x for YOLO11, and YOLOv12n, YOLOv12s, YOLOv12m, and YOLOv12l for YOLOv12 to evaluate their efficiency in detecting the fruitlets in commercial orchards. Total model parameters, GFLOPs, and convolutional layers were evaluated using Equations (3)–(5), respectively.

$$Precision = \frac{TP}{\mathrm{TP} + \mathrm{FP}} \tag{1}$$

$$Recall = \frac{TP}{\mathrm{TP} + \mathrm{FN}} \tag{2}$$

$$Parameters_{Model} = \text{Total trainable weights and biases} \tag{3}$$

$$GFLOPs = \frac{\textit{Total floating point operations per image}}{10^9} \tag{4}$$

$$Layers_{CONVOLUTIONAL} = \text{Total number of convolutional layers in the model} \tag{5}$$

Additionally, the assessment involved analyzing pre-processing, inference, and post-processing speeds for each model configuration, as these metrics are critical for real-time object detection systems. Pre-processing speed determines how quickly a model can prepare images for detection, inference speed measures the time taken to identify objects within images, and post-processing speed reflects how swiftly the model finalizes the outputs after detection. All these stages are essential for efficient operation in agricultural applications, as timely and accurate detection directly impacts decision-making and resource management. Therefore, this evaluation of computational efficiency is crucial for advancing practical agricultural technology solutions.

### 2.4. In-field counting validation

Upon evaluating the performance metrics of each YOLOv8, YOLOv9, YOLOv10, YOLO11, and YOLOv12 configuration for detecting fruitlets, the most effective model from each version was selected based on the highest accuracy achieved at mAP@50. These top-performing models from YOLOv8, YOLOv9, YOLOv10, YOLO11, YOLOv12 were further validated to assess their detection capabilities in a real commercial orchard setting across four distinct apple varieties in Washington State. This validation process utilized images collected both with a smartphone and machine vision sensor. Images of SciFresh, SciLate, Cosmic Crisp, and Honeycrisp apples were captured using the smartphone. For the evaluation, a total of 128 images (32 per variety) were analyzed to assess the models' counting accuracy before the thinning process.

Subsequently, an additional set of images was used for further validation. This included 32 images of SciFresh apples taken with an Intel RealSense machine vision camera. Notably, while the Intel RealSense camera images of SciFresh and SciLate apples served as the training dataset, validation was performed on varieties: Cosmic Crisp and Honeycrisp, which were not included in the training phase. For each image, a count of all visible and occluded apple fruitlets was manually performed directly in the field. This manual counting provided a precise ground truth for each image sample across the different apple varieties, ensuring robust validation of the models' performance.

In our study assessing the performance of YOLO models for detecting fruitlets in commercial orchards, Root Mean Square Error (RMSE) and Mean Absolute Error (MAE) were employed as crucial metrics to quantify the accuracy of fruit counts predicted by the models compared to manually counted ground truths.

RMSE was calculated using Equation (6). Here, predicted_i denotes the model-predicted fruitlet count, actual_i represents the manual ground-truth count for the i-th image, and $n$ is the total number of evaluated images. This measure computes the square root of the average of the squared differences between the predicted and actual counts, thereby emphasizing larger errors and highlighting significant deviations in model performance.

Likewise, MAE was determined using Equation (7). In this equation, predicted_i and actual_i retain their previous definitions, with MAE calculating the average of the absolute differences between the predicted and actual counts. In contrast to RMSE, MAE is less sensitive to large errors and provides a straightforward indication of the average error magnitude per image, thereby offering an intuitive measure of prediction accuracy across the dataset.

$$RMSE = \sqrt{\left(\frac{1}{n}\sum (predicted_i - actual_i)\hat{}2\right)} \tag{6}$$

$$MAPE = \frac{1}{n}\sum_{t=1}^{n} |(predicted_i - actual_i)| \tag{7}$$

### 2.5. Architectural robustness to cross-sensor domain shifts

In this study, training data were exclusively collected using the Intel RealSense D435i RGB camera, while in-field validation was performed on images acquired using an iPhone 14 Pro Max. These sensors differ significantly in resolution, image tone, color calibration, and lens characteristics, introducing a cross-sensor domain shift that poses a challenge to model generalization. Although no explicit domain adaptation techniques (e.g., adversarial training or feature alignment) were applied, the YOLO variants evaluated in this study incorporate architectural elements that inherently enhance robustness across sensor domains. For instance, YOLO11 includes the C2PSA and C3K2 modules, which refine spatial attention and reduce feature degradation due to background variability or lighting variations [3,62]. YOLOv12 introduces Area Attention and R-ELAN, enabling stronger global and local context modeling [62]. YOLOv9 integrates GELAN and PGI, both of which enhance multi-scale feature aggregation and stable learning dynamics critical for resolving cross-sensor differences in illumination and object appearance. To further promote generalization, our training dataset included substantial variability in lighting (direct sun, cloud cover, partial shade), viewing angles (30°–90° zenith tilt, 0°–180° azimuth), and occlusion levels. This diverse training distribution implicitly enhances model robustness against unseen sensor domains, such as smartphone-based imagery, without fine-tuning.

## 3. Results and discussion

Fig. 4 illustrates the detection outcomes of the fastest variants (nano) of YOLOv8, YOLOv9, YOLOv10, YOLO11, and YOLOv12 on two representative images captured in commercial apple orchards. In the left image, the original image was displayed at the top, with the predictions from each model arranged sequentially below it. In this example, a green arrow on the upper portion of the image indicated an apple fruitlet partially visible in a complex scene, while another green arrow on the lower portion pointed to an apple fruitlet under severe occlusion. The green arrows denoted instances where the model correctly detected the fruitlets, whereas a red arrow at the same spatial location highlighted a missed detection. Specifically, YOLOv8 correctly identified the former but failed to detect the latter. In contrast, YOLOv9 missed the former yet successfully detected the latter. YOLOv10, however, failed to detect either fruitlet in this scenario. YOLO11 detected the former but again failed on the latter, and YOLOv12 also managed to identify only the former, which highlights the variability in detection performance across the different YOLO architectures.

In the right image, the original image was again presented at the top, followed by the predictions from the different YOLO models. A green dotted circle marked a region containing three apple fruitlets, with one fruitlet partially obscured by two others in the foreground. In this region, YOLOv8 detected only two fruitlets, as evidenced by the dotted circle, while YOLOv9 and YOLOv10 similarly failed to detect all three. Notably, YOLO11 successfully identified all fruitlets within the occluded region, demonstrating improved robustness under challenging conditions. Conversely, YOLOv12, despite being the latest iteration, failed to detect the partially occluded fruitlet in that area. These results underscore the significant variability in detection performance across the YOLO variants, particularly in complex and occluded scenes. Overall, the findings revealed that while YOLO11 exhibited superior detection performance in certain challenging scenarios, models such as YOLOv10 and YOLOv12 demonstrated limitations. This variability emphasizes the need for further refinement in model architecture and training strategies to enhance robustness and accuracy in real-world agricultural applications.

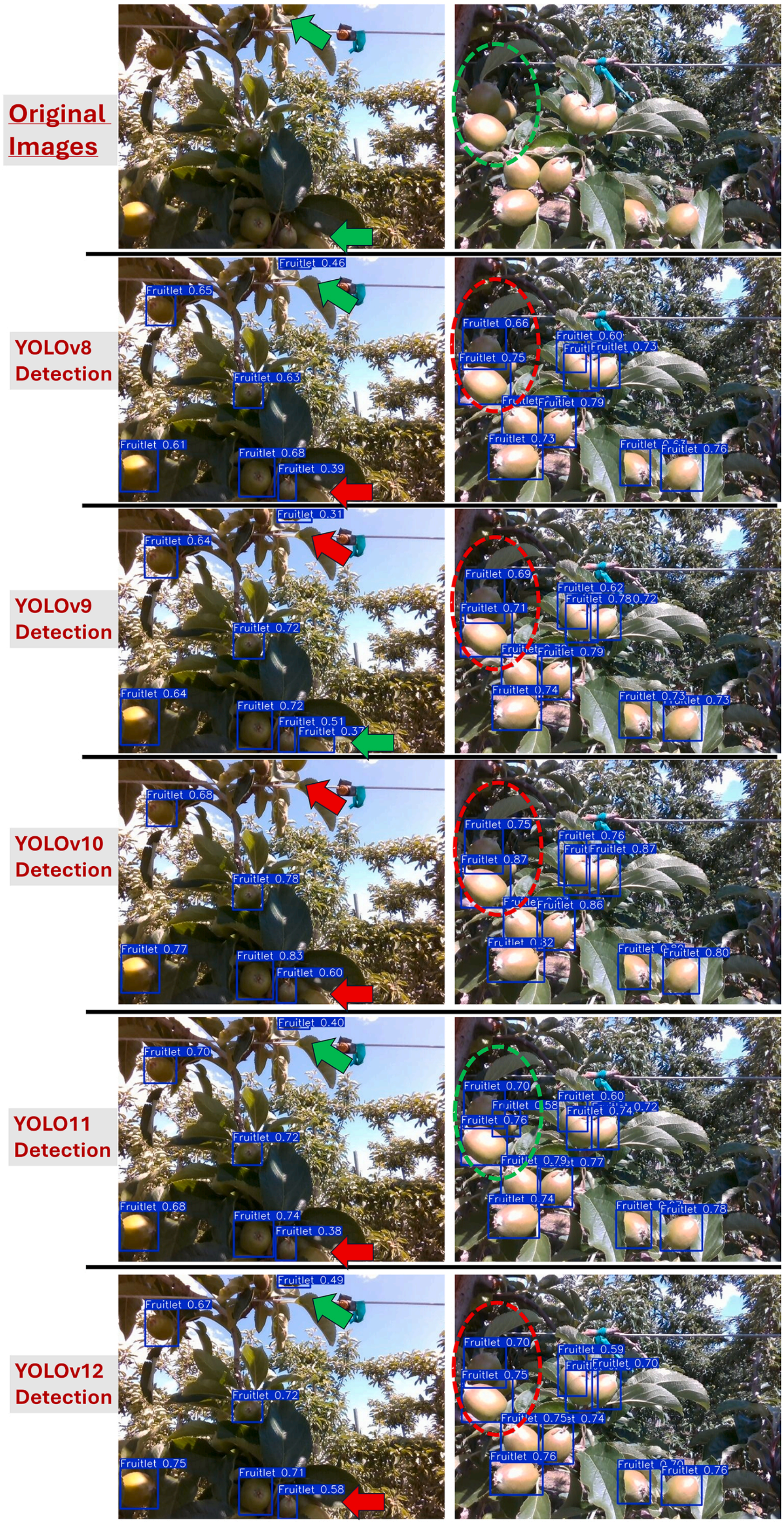

**Fig. 4.** Illustration of Object Detection (Green Fruitlet) results with YOLOv12, YOLO11, YOLOv10, YOLOv9, and YOLOv8 models. The figure presents original images (left) and predictions (right) with green and red arrows indicating correct and missed detections.

### *3.1. Assessment of detection accuracy: Precision and recall metrics*

Among all the models compared in this study across the configurations of YOLOv8 to YOLOv12, YOLOv10x achieved the highest precision score of 0.908, while YOLOv12l attained the highest recall score of 0.900. Within each model family, performance differences were evident. In the YOLOv12 family which comprised four fully trained variants (YOLOv12n, YOLOv12s, YOLOv12m, and YOLOv12l, noting that YOLOv12x could not be completely trained due to memory constraints) YOLOv12l demonstrated the highest precision at 0.87. In the YOLO11 family, among the evaluated variants, YOLO11l achieved the highest precision with a score of 0.899. Similarly, within the YOLOv9 series, the variant YOLOv9 GELAN-c achieved the highest precision score of 0.903 among the six configurations tested, while in the YOLOv8 family comprising five configurations, YOLOv8m obtained the highest precision, reaching a value of 0.897. These findings reflect varying performance across different YOLO architectures and underscore the critical importance of model selection and training optimization for specific detection tasks. Detailed precision and recall values for all the evaluated models are presented in Table 2, providing a comprehensive overview of detection performance across the various YOLO families. Overall, these results highlight the trade-offs between precision and recall inherent in different model architectures and emphasize the need for further optimization to achieve balanced performance in real-time object detection applications.

It is critical to analyze the architectural principles contributing to these observed differences. For instance, the superior precision of YOLOv10x (0.908) may be attributed to its dual assignment strategy, which reduces label redundancy during training by integrating both one-to-one and one-to-many matching. This enhances label quality and spatial discrimination. In contrast, YOLOv9 GELAN-c's high precision (0.903) benefits from the GELAN backbone, which facilitates multi-scale aggregation with lower information loss. YOLO11's consistently strong recall performance, particularly in YOLO11l (0.883), is likely enhanced by the introduction of the C3K2 module, which improves gradient flow and feature preservation across scales, and the inclusion of the C2PSA attention block, which improves spatial localization of overlapping fruitlets. However, YOLOv12l, despite incorporating advanced R-ELAN and Area Attention ($A^2$) modules for deeper context-aware feature fusion, demonstrated marginal improvements in recall but not consistently in precision, which may be attributed to over-attention mechanisms that can suppress fine-scale details in clustered fruitlets. These insights suggest that while recent modules improve general detection robustness, they may introduce trade-offs between spatial accuracy and false positives, especially under heavy occlusion and background clutter.

### *3.2. Evaluation of detection consistency: Mean average precision at IoU = 0.50*

The assessment of mean Average Precision at an Intersection over Union (IoU) threshold of 0.50 (mAP@50) across various YOLO model configurations in Fig. 5 provides profound insights into their efficacy in detecting fruitlets within agricultural settings. Among all evaluated configurations, YOLOv9 models, particularly YOLOv9 GELAN-e and YOLOv9 GELAN-base, stand out, both achieving the highest mAP@50 scores of 0.935. These scores, which exceed those of all configurations of YOLOv8, YOLOv10, YOLO11 and YOLOv12, underscore YOLOv9's superior detection accuracy. Furthermore, as depicted in Fig. 6, all configurations of the YOLOv9 family consistently achieved a higher mAP@50 compared to the other model families evaluated in this study.

**Table 2**
Precision and Recall for YOLO Configurations (YOLOv8, YOLOv9, YOLOv10, YOLO11, and YOLOv12) used in Fruitlet Detection. For F1-score metrics visualization, refer to Fig. 7.

| Configuration | Precision | Recall |
|---|---|---|
| YOLOv8n | 0.840 | 0.883 |
| YOLOv8s | 0.883 | 0.864 |
| YOLOv8m | 0.897 | 0.858 |
| YOLOv8l | 0.877 | 0.852 |
| YOLOv8x | 0.870 | 0.865 |
| YOLOv9 GELAN-e | 0.875 | 0.891 |
| YOLOv9 GELAN-c | 0.903 | 0.872 |
| YOLOv9 GELAN-s | 0.877 | 0.881 |
| YOLOv9 GELAN-t | 0.873 | 0.864 |
| YOLOv9 GELAN-m | 0.866 | 0.899 |
| YOLOv9 GELAN-base | 0.881 | 0.895 |
| YOLOv10n | 0.891 | 0.862 |
| YOLOv10s | 0.839 | 0.872 |
| YOLOv10m | 0.877 | 0.864 |
| YOLOv10b | 0.898 | 0.870 |
| YOLOv10l | 0.842 | 0.872 |
| YOLOv10x | **0.908** | 0.846 |
| YOLO11n | 0.897 | 0.868 |
| YOLO11s | 0.881 | 0.883 |
| YOLO11m | 0.858 | 0.897 |
| YOLO11l | 0.899 | 0.883 |
| YOLO11x | 0.891 | 0.866 |
| YOLOv12n | 0.864 | 0.881 |
| YOLOv12s | 0.880 | 0.840 |
| YOLOv12m | 0.860 | 0.883 |
| YOLOv12l | 0.871 | **0.900** |

The newly included YOLOv12 and YOLO11 series introduce impressive scores as well: YOLOv12l records a mAP@50 of 0.931, YOLOv12m at 0.928. YOLO11n records a mAP@50 of 0.926, YOLO11s at 0.933, YOLO11m at 0.924, YOLO11l at 0.932, and YOLO11x at 0.922, indicating enhanced detection capabilities across these newer configurations. Within the YOLOv10 series, YOLOv10n leads with a mAP@50 of 0.921, closely followed by YOLOv10b and YOLOv10m, which record scores of 0.919 and 0.917, respectively. Despite their strong performance, these figures remain slightly below the peak values presented by YOLOv9, indicating that specific enhancements in YOLOv9 have likely boosted accuracy. In contrast, the YOLOv8 configurations show a wider performance spectrum. YOLOv8s achieves the highest mAP@50 within its group at 0.924, nearly matching the top-performing YOLOv10 configurations. Nonetheless, configurations such as YOLOv8l and YOLOv8x achieved lower scores of 0.912 and 0.914, respectively. This variation may be attributed to overparameterization of the models for the given task. Considering the dataset's simplicity and fewer categories relative to the more complex COCO dataset, the extensive parameters in these models might be excessive and potentially hinder optimal performance.

### *3.3. Analysis of computational efficiency: Image processing speed*

In assessing computational efficiency, particularly image processing speeds, YOLO11 emerged as the top performer, achieving remarkably low inference times across its variants, with an exceptional time of just 2.4 ms. Moreover, the pre-processing times for YOLO11 were notably quick, registering 0.1 ms for all variants, except for YOLO11l, which still performed impressively at 0.2 ms. This rapid pre-processing facilitates faster preparation of images for subsequent detection phases. Despite

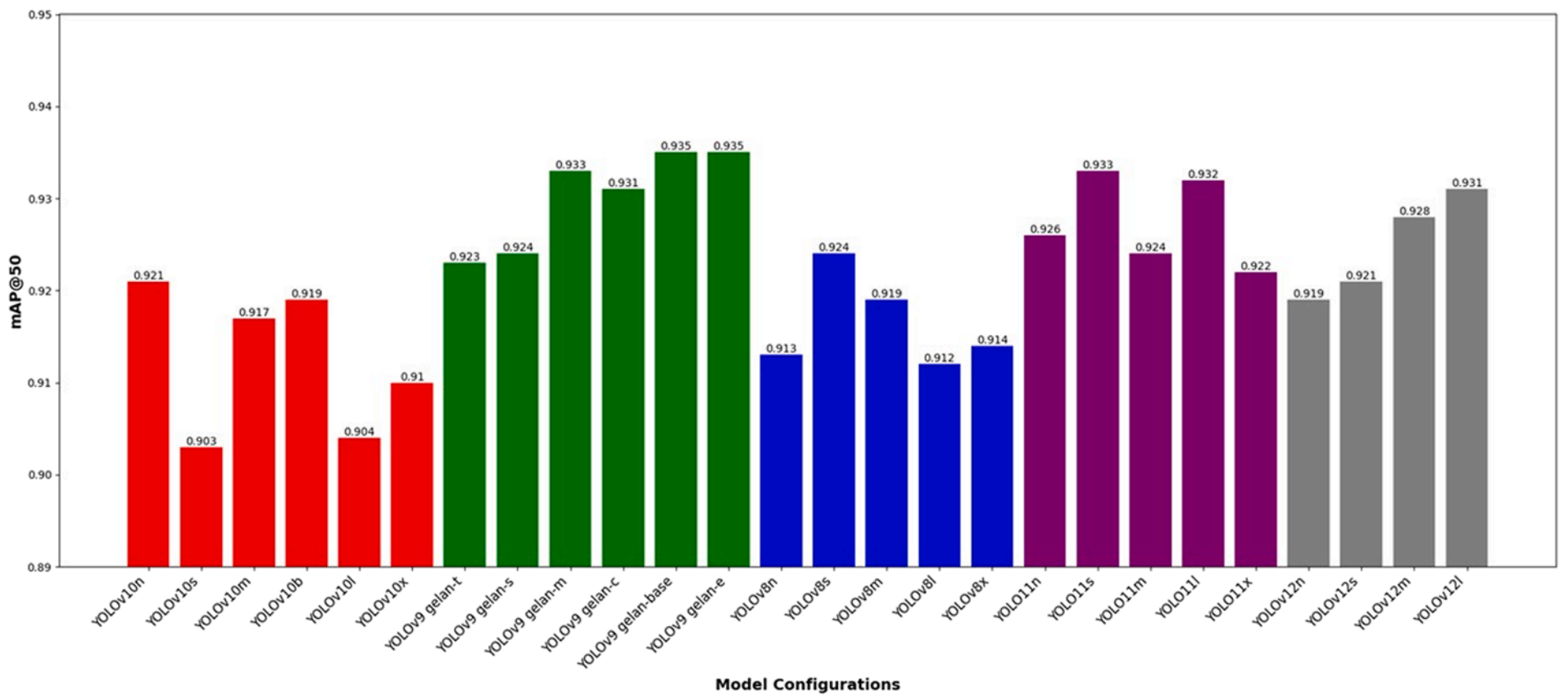

**Fig. 5.** Bar diagram showing mAP@50 scores for all Configurations of YOLOv8, YOLOv9, YOLOv10, YOLO11, and YOLOv12 for fruitlet detection in commercial orchards.

YOLO11's superior performance, YOLOv12n, YOLOv12m, and YOLOv8x were also notable within their series for achieving the fastest pre-processing speed at merely 0.8, 0.3, and 0.9 ms respectively. While pre-processing speeds are expected to be consistent across models, the discrepancies observed, particularly in the speed of YOLOv12n, YOLOv12m, and YOLOv8x, are likely due to random variations rather than fundamental differences in model architecture.

Expanding on the theme of processing efficiency, YOLO11n configurations demonstrated excellence in inference speed. YOLO11n, in particular, recorded a time of 2.4 ms, significantly outpacing the fastest models from the YOLOv12, YOLOv10, YOLOv9 and YOLOv8 series. In comparison, the quickest YOLOv9 model, YOLOv9 GELAN-s, logged an inference time of 11.5 ms, and the leading YOLOv10 model, YOLOv10s, achieved 5.5 ms. These results underscore YOLO11n's superior capability in rapid image processing, highlighting its robustness and suitability for scenarios that demand high-speed, accurate object detection. This analysis reveals that while the YOLO11 series leads in low-latency performance. The detailed performance metrics for each model configuration across the YOLOv8, YOLOv9, YOLOv10, and YOLO11 series are systematically presented in Table 3.

### 3.4. *Field validation of counting accuracy: RMSE and MAE metrics*

For the counting validation performed on four apple varieties using images collected with an iPhone 14 Pro Max smartphone, the top-performing configurations from each YOLO version (YOLOv8n, YOLOv9 GELAN-e, YOLOv10n, YOLO11n, and YOLOv12n) were rigorously assessed for green fruitlet counting accuracy. Among these, YOLO11n demonstrated exceptional performance by consistently yielding lower error metrics compared to its predecessors. In particular, for Honeycrisp apples, YOLO11n recorded an RMSE of 4.51 and an MAE of 4.07, marking a significant improvement in detection accuracy and reliability. Similarly, for Cosmic Crisp apples, the configuration achieved an RMSE of 4.59 and an MAE of 3.98. These results underscore the model's robustness in handling subtle variations in fruit appearance and background complexity, indicating that YOLO11n is highly effective in accurately detecting and counting green fruitlet under challenging in-field conditions.

In contrast, for the SciLate variety, YOLO11n produced an RMSE of 4.83 and a notably higher MAE of 7.73, while for SciFresh apples it achieved an RMSE of 4.96 and an MAE of 3.85. These differences highlight the impact of fruit variety and environmental factors on model performance, suggesting that detection accuracy can vary with the inherent complexity of the orchard scene. Fig. 7a provides a comprehensive comparison of RMSE (left) and MAE (right) for green fruitlet detection across the evaluated YOLO configurations, effectively illustrating these performance variations. Overall, the quantitative findings indicate that YOLO11n offers superior accuracy and precision in in-field counting validation, demonstrating its advanced capability to reliably detect and count green fruitlets. These insights pave the way for further enhancements in automated fruit detection and precision agriculture applications.

It is noteworthy that the counting validation was conducted on images collected using an iPhone 14 Pro Max, which were not part of the training dataset; the models were exclusively trained on images captured by an Intel RealSense camera. Despite this domain gap, the top-performing configurations, especially YOLO11n and YOLOv9 GELAN-e, demonstrated robust counting accuracy across four apple varieties. In particular, YOLOv9 GELAN-e excelled with Honeycrisp apples by achieving an RMSE of 4.89 and an MAE of 4.14, while for Cosmic Crisp, it registered an RMSE of 4.77 and an MAE of 4.04. For the SciLate variety, the model produced an RMSE of 5.54 and an MAE of 4.16, and for SciFresh, it achieved an RMSE of 5.37 and an MAE of 4.23. These performance metrics indicate that YOLOv9 GELAN-e is among the top-performing models, demonstrating highly competitive accuracy in precision fruit counting under these novel imaging conditions when compared to the YOLOv8, YOLOv10, YOLOv11, and YOLOv12 configurations.

The results clearly illustrate the models' ability to generalize to new data, even when the images are acquired with different sensors. Fig. 7b provides a detailed visual distribution of RMSE and MAE values, illustrating the in-field counting validation for green fruitlet detection using iPhone 14 images. This evaluation not only confirmed the strong performance of YOLO11n but also underscored the competitive robustness of YOLOv9 GELAN-e under challenging conditions. The observed discrepancies between model configurations emphasize the need for further exploration into domain adaptation and cross-sensor generalization techniques. Ultimately, these findings highlight the potential for training models with high-quality machine vision sensor data to produce models that maintain high accuracy when deployed with more accessible smartphone cameras, thereby enhancing practical applications in precision agriculture.

In the evaluation of images captured by standard machine vision cameras (Intel RealSense) for apple counting, the updated results highlight the exceptional accuracy of the YOLO11n configuration. Specifically, YOLO11n demonstrated remarkable performance in counting SciFresh apples, achieving an RMSE of 3.06 and an MAE of 2.33. This outperforms the previously noted accuracy of the YOLOv9 GELAN-e configuration, which topped earlier assessments with an RMSE of 3.11 and an MAE of 2.46 for the same variety. Fig. 7b and c illustrate the distribution of RMSE and MAE across multiple configurations and apple varieties. Specifically, Fig. 7b details results for SciLate, SciFresh,

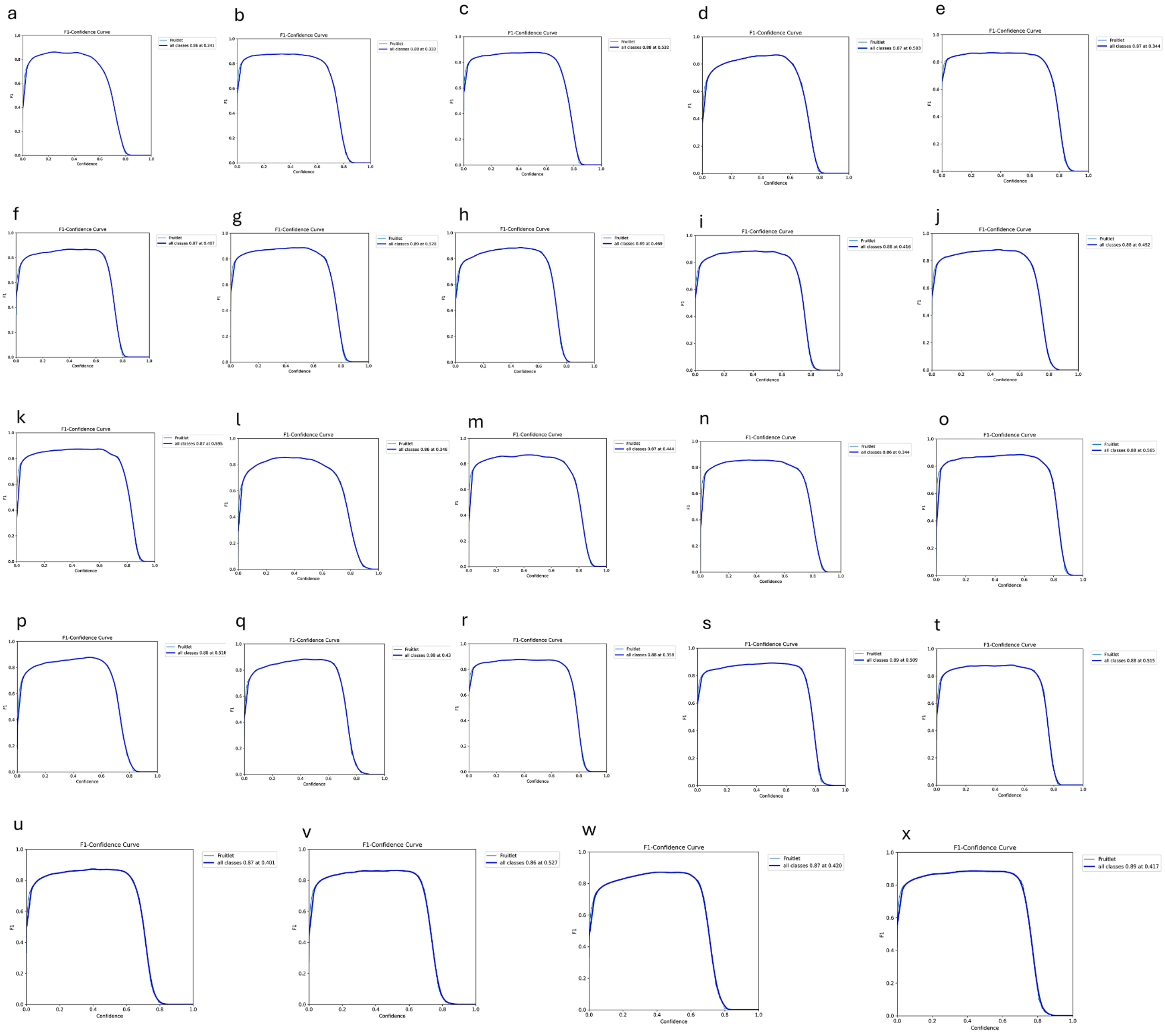

**Fig. 6.** F1-scores of YOLO models (YOLOv8, YOLOv9, YOLOv10, YOLO11, and YOLOv12) for fruitlet detection. (**a**)–(**e**) illustrate F1 scores achieved with various YOLOv8 configurations (YOLOv8n, YOLOv8s, YOLOv8m, YOLOv8l, YOLOv8x). (**f**)–(**k**) same with YOLOv9 configurations (YOLOv9 GELAN-t, GELAN-s, GELAN-m, GELAN-c, GELAN-base, GELAN-e). (**l**)–(**q**) with YOLOv10 configurations; (r)–(v) with YOLO11 configurations. (**w**)–(**z**) with YOLOv12 configurations (YOLOv12n, YOLOv12s, YOLOv12m, YOLOv12l), highlighting comparative detection accuracy of these models in complex orchard environments.

Honeycrisp, Cosmic Crisp apples captured with an iPhone 14 Pro Max, while Fig. 7c focuses on the performance for SciFresh apples captured using Intel RealSense cameras. These findings highlight the variable performance of each model and underscore the need for larger, more diverse training datasets.

### 3.5. Evaluation of cross-sensor domain generalization

To evaluate the YOLO models' capacity for domain generalization, we conducted in-field validation using 128 images captured with an Apple iPhone 14 Pro Max, a sensor distinct from the Intel RealSense camera used for training. This evaluation was designed to simulate practical deployment conditions where hardware variability may introduce domain shift. Among the tested configurations, YOLO11n and YOLOv9 GELAN-e exhibited the strongest resilience to cross-sensor differences. Specifically, YOLO11n achieved the lowest RMSE and MAE across three out of four apple varieties, including Honeycrisp and Cosmic Crisp, which were not seen during training. These results suggest that the C2PSA and C3K2 modules effectively preserve spatial features and semantic consistency even when the image domain shifts significantly. Likewise, YOLOv9 GELAN-e maintained high accuracy, particularly on Cosmic Crisp images (RMSE: 4.77; MAE: 4.04), likely due to the multi-scale feature aggregation capabilities of GELAN and the adaptive gradient optimization introduced by PGI. Importantly, the models were not fine-tuned or retrained on smartphone images. Their generalization ability reflects the impact of both architectural innovations and the deliberately diverse environmental conditions used during training. These findings emphasize the potential of modern YOLO architectures to be deployed across heterogeneous imaging platforms in real-world orchard robotics, where sensor uniformity cannot always be guaranteed.

**Table 3**
Performance of pre-processing, inference, and post-processing stages.

| YOLO Versions | YOLO Configurations | Pre-processing(ms) | Inference(ms) | Post-processing(ms) |
|---|---|---|---|---|
| YOLOv8 | YOLOv8n | 1.3 | 4.1 | 2.3 |
| | YOLOv8s | 1.3 | 6.4 | 2.3 |
| | YOLOv8m | 1.3 | 11.2 | 2.1 |
| | YOLOv8l | 1.2 | 18.7 | 2.2 |
| | YOLOv8x | 0.9 | 24.8 | 2.3 |
| YOLOv9 | YOLOv9 GELAN-t | 1.3 | 14.1 | 2.2 |
| | YOLOv9 GELAN-s | 1.3 | 11.5 | 2.2 |
| | YOLOv9 GELANn-m | 1.3 | 14.0 | 2.1 |
| | YOLOv9 GELAN-c | 1.3 | 17.0 | 2.0 |
| | YOLOv9 GELAN-base | 1.2 | 17.2 | 2.0 |
| | YOLOv9 GELAN-e | 1.1 | 33.5 | 1.9 |
| YOLOv10 | YOLOv10n | 1.4 | 5.5 | 1.6 |
| | YOLOv10s | 1.3 | 7.7 | 1.6 |
| | YOLOv10m | 1.3 | 13.0 | 1.6 |
| | YOLOv10b | 1.3 | 16.7 | 1.5 |
| | YOLOv10l | 1.4 | 19.6 | 1.5 |
| | YOLOv10x | 1.2 | 26.5 | 1.5 |
| YOLO11 | YOLO11n | **0.1** | **2.4** | 2.2 |
| | YOLO11s | **0.1** | 5.0 | 2.5 |
| | YOLO11m | **0.1** | 11.9 | 0.6 |
| | YOLO11l | 0.2 | 14.6 | 1.2 |
| | YOLO11x | **0.1** | 26.3 | 0.6 |
| YOLOv12 | YOLOv12n | 0.8 | 4.6 | 0.7 |
| | YOLOv12s | 3.8 | 9.6 | 1.0 |
| | YOLOv12m | 0.3 | 15.9 | **0.5** |
| | YOLOv12l | 0.6 | 19.2 | **0.5** |

Note: Pre-processing: Image preparation (scaling, normalization). Inference: Core detection phase. Post-processing: Output refinement including Non-Maximum Suppression. Bold values indicate best performance in each category.

### *3.6. Architectural innovation analysis and its impact on performance*

While previous sections detailed the numerical performance of various YOLO configurations, a deeper investigation into the architectural design is essential to understand the drivers behind such results. Each YOLO version integrates distinct innovations that impact precision, inference speed, and mean Average Precision (mAP@50). YOLO11n, which achieved the fastest inference speed (2.4 ms), integrates the C3K2 block and C2PSA attention as key design components [3,63]. The C3K2 block, an enhancement over the traditional C2f, reduces channel expansion and computational depth while maintaining feature reuse, enabling faster forward propagation [64,65]. Meanwhile, the C2PSA (Convolutional Block with Parallel Spatial Attention) refines spatially salient features with minimal latency overhead [66–68]. Together, these lightweight yet expressive modules ensure high-speed detection without compromising accuracy.

In contrast, YOLOv9 models (especially GELAN-e and GELAN-base) excelled in mAP@50 (0.935), primarily due to their use of the GELAN and PGI. GELAN enables dynamic multi-scale feature aggregation [69,70], effectively preserving high-resolution details that are critical for fruitlet detection in occluded and cluttered orchard environments. PGI ensures stable backpropagation by adapting the learning dynamics to emphasize critical gradient paths [3], which reduces convergence errors and improves precision in bounding box regression. YOLOv10's high precision is supported by its dual assignment strategy, which removes the need for non-maximum suppression by assigning detection targets through both one-to-one and one-to-many matching [71,72]. This innovation minimizes overlap ambiguity and enhances object localization. Furthermore, its rank-guided block optimizes channel-wise feature prioritization, boosting precision while maintaining speed [60].

YOLOv12, although equipped with advanced Area Attention ($A^2$) and R-ELAN modules [3], did not consistently outperform YOLOv9 in mAP. The Area Attention dynamically adjusts the receptive field to enhance global context understanding. However, it may dilute the fine-grained features that are critical for small object detection when applied too broadly [65]. R-ELAN stabilizes deep gradient flow but adds minor computational overhead [62].

### *3.7. Discussion*

Fig. 8 presents several examples of green fruitlet detection on smartphone-captured images of two apple varieties, Honeycrisp and Cosmic Crisp, which were not part of the training set for any of the YOLO models evaluated. The figure is divided into five panels, each illustrating the detection performance of different YOLO configurations. For instance, Fig. 8a depicts YOLOv8n's performance on Cosmic Crisp apples; despite the apples being clearly visible, albeit partially occluded as highlighted by red dotted regions, YOLOv8n failed to detect the apple in the lower region, where approximately 90 % of the fruit was visible. In Fig. 8b, the YOLOv9 GELAN-base model detected most of the green fruitlets in the frame, yet it missed one apple, as indicated by red dotted circles. The apple in the upper region was severely occluded and blended closely with the background, which likely contributed to the detection failure. Similarly, Fig. 8c illustrates YOLOv10n's performance on a densely clustered scene in Cosmic Crisp apples, where one apple, marked with a red-dotted circle, was not detected.

Furthermore, Fig. 8d shows the detection results of YOLO11n on Cosmic Crisp apples, where the model missed two apples due to occlusion. In the final panel, Fig. 8e, the YOLOv12 configuration is evaluated for green fruitlet detection; here, the red dotted circles indicate apples that were not detected. Although the model managed to detect some of the smaller, occluded fruitlets in complex scenes, YOLOv12 failed to detect a clearly visible fruitlet in the upper red dotted region. These diverse outcomes underscore the variable detection capabilities of the different YOLO models when confronted with unseen data acquired from a smartphone. Overall, the results from this study, which involved 1149 images, suggest that while current models show promise, there is a critical need to expand and diversify training datasets and to utilize more computationally intensive training environments. Such improvements would likely enhance model accuracy and generalizability in real-world fruit detection applications.

Recent work on green fruit detection has addressed challenges in complex orchard environments through diverse approaches. Liu et al. [73] improved segmentation by replacing the FPN with Residual FPN (RFPN) and adding a double-layer Convolutional Block Attention

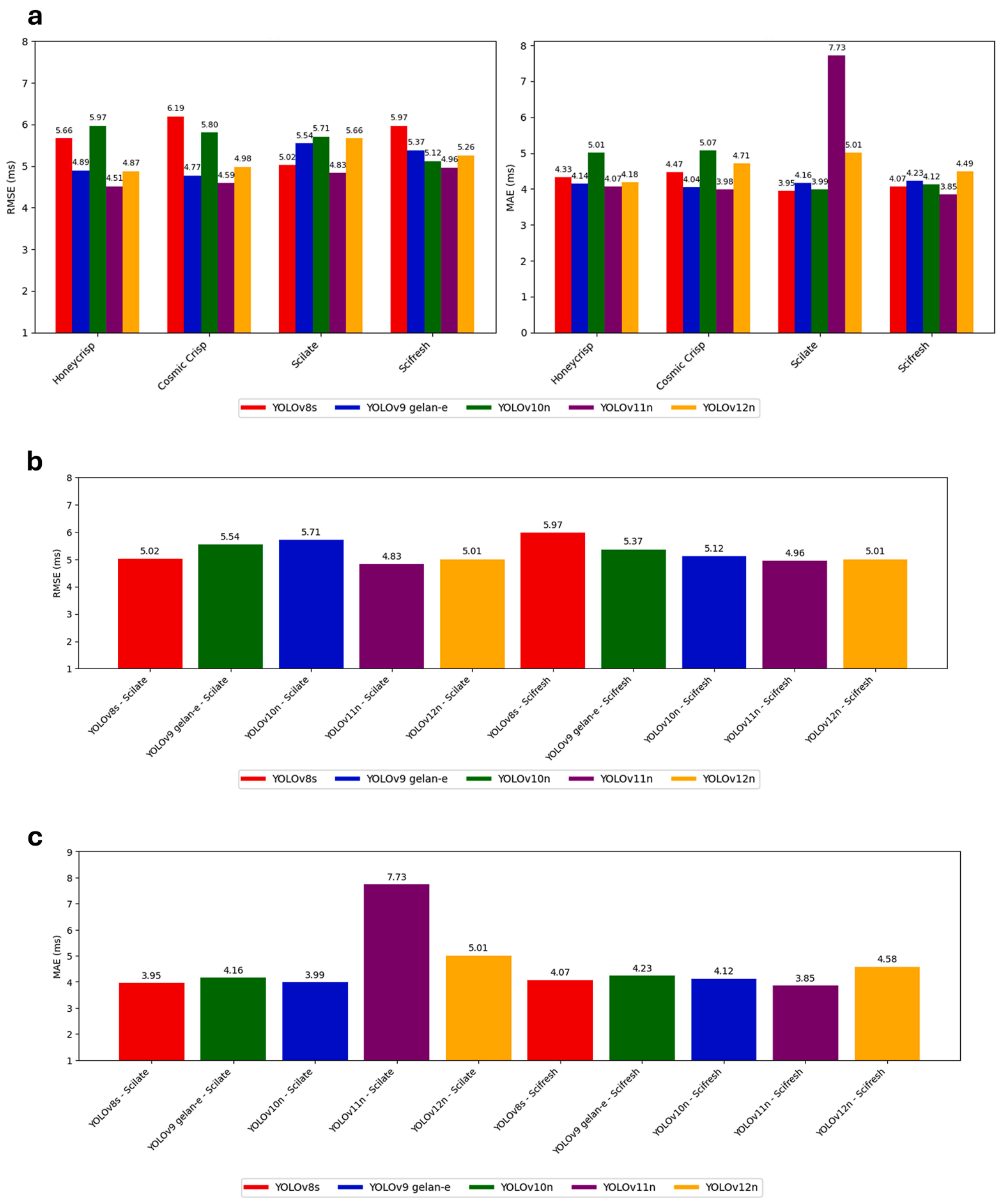

**Fig. 7.** Illustration of in-field validation of YOLOv8, YOLOv9, YOLOv10, YOLO11 and YOLOv12 configurations in detecting fruitlets. (**a**) RMSE and MAE for in-field counting validation for apple fruitlet detection using iPhone 14 Pro Max Images: Bar diagram on the left side shows the comparison of RMSE while the bar diagram on right shows the MAE for counting validation of immature green fruitlets. (**b**) RMSE and MAE for apple fruitlet detection using a smartphone (iPhone 14 Pro Max). (**c**) MAE for apple fruitlet detection across YOLOv8s, YOLOv9 GELAN-e, YOLOv10n, YOLO11n, and YOLOv12n assessed with Intel RealSense camera.

Module (CBAM), achieving a mAP of 85.3 % for occluded fruits. Jia et al. [74] proposed Fast-FDM, an anchor-free Foveabox model with EfficientNetV2-S, BiFPN, and Adaptive Training Sample Selection (ATSS), achieving 62.3 % mAP with reduced costs. Hussain et al. [75] applied Mask R–CNN with orientation estimation, attaining APs of 83.4 % overall and 91.3 % for larger masks. Sun et al. [76] introduced GHFormer-Net, which integrates gradient harmonizing mechanisms on a RetinaNet-PVTv2 backbone, achieving 85.2 % AP under low light orchard conditions. In contrast, YOLO11n offers 2.4 ms inference speed with strong precision/recall, excelling even on smartphone images, making it practical for real-time precision agriculture.

## 4. Conclusion and future suggestions

In this study, we conducted a comprehensive evaluation of various configurations of state-of-the-art YOLO object detection models (YOLOv8, YOLOv9, YOLOv10, YOLO11 and YOLOv12) to assess their effectiveness in detecting green fruitlets before thinning in complex orchard environments using different sensors and conditions. The key findings and future work of this study are summarized as follows:

(1) Model-Specific Performance. YOLOv10x and YOLOv9 GELAN-c demonstrated high performance in precision and recall metrics, respectively. YOLOv10x achieved the highest precision of 0.908, followed by YOLOv9 GELAN-c at 0.903. These results indicate superior accuracy in correctly identifying fruitlets with minimal false positives. Meanwhile, YOLOv12l achieved the highest recall of 0.900, indicating its strong capability in detecting almost all fruitlets in the images.

(2) Accuracy Across Configurations. For mean Average Precision at a 50 % Intersection over Union (mAP@50), YOLOv9 configurations particularly YOLOv9 GELAN-base and YOLOv9 GELAN-e achieved the highest mAP@50 of 0.935, demonstrating good robustness in agricultural object detection. Furthermore, YOLO11s and YOLOv12l performed comparably well, with mAP@50 values of 0.933 and 0.931, respectively, demonstrating the efficacy of these advanced YOLO configurations for precise object detection in agricultural settings.

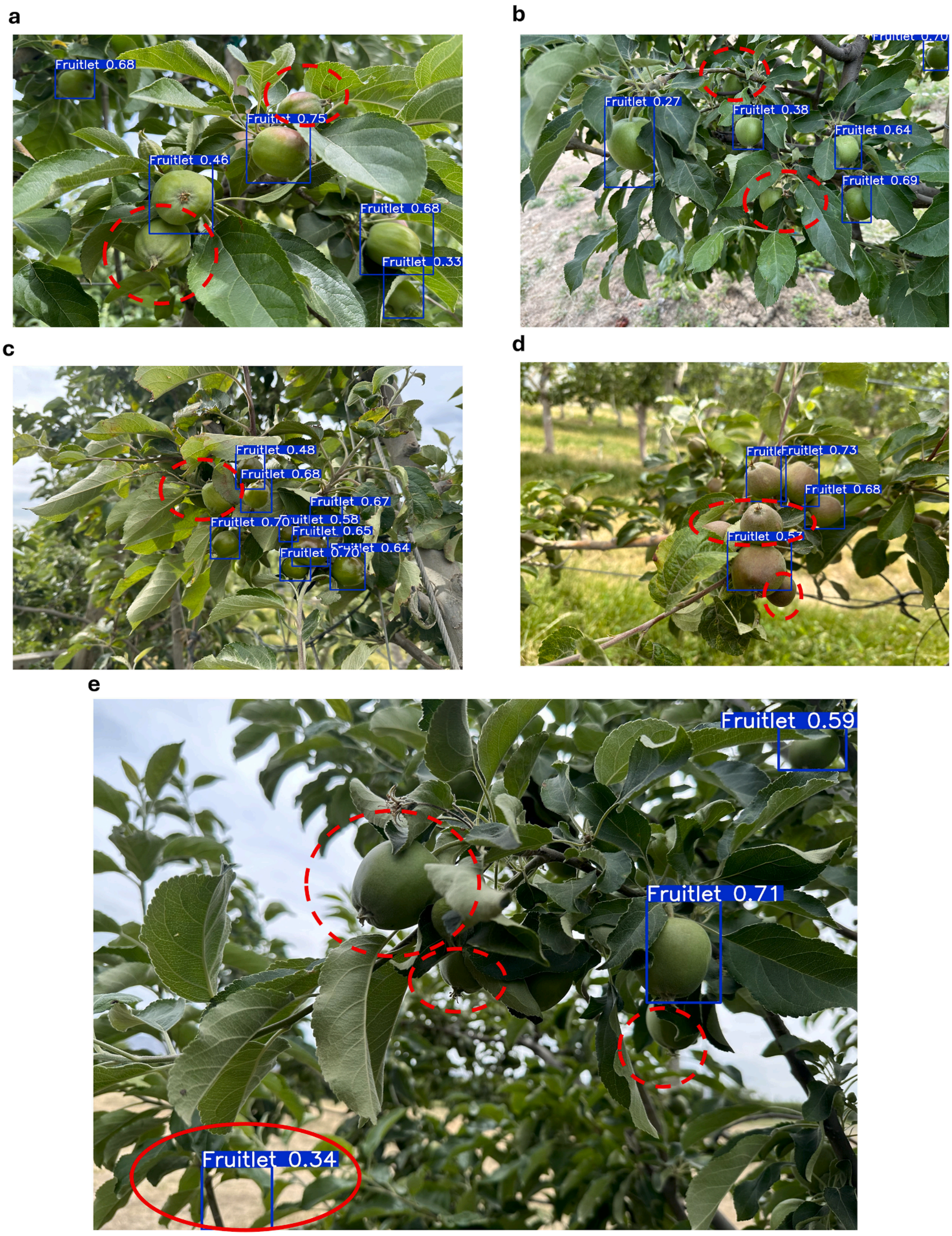

**Fig. 8.** Examples of results for apple fruitlet detection using iPhone 14-based images collected from Honeycrisp and Cosmic Crisp varieties, which were not used during the model training process. (**a**) YOLOv8n missed nearly 10 % of the visible fruitlets in a partially occluded region (red-dotted area). (**b**) YOLOv9 GELAN-base detected most of the apples but missed one in both upper and lower occluded areas. (**c**) YOLOv10n failed to detect a clearly visible apple in a densely clustered scene. (**d**) YOLO11n missed two apples due to occlusion. (**e**) YOLOv12 failed to detect a clearly visible fruitlet despite partial detection of smaller, occluded fruitlets.

(3) Counting Validation. The YOLO11n model demonstrated high precision, achieving the lowest RMSE and MAE across all tested varieties, which underscores its robustness in apple fruitlet detection. Specifically, in iPhone 14 Pro Max-captured images, YOLO11n achieved the best RMSE for each variety: 4.51 for Honeycrisp, 4.59 for Cosmic Crisp, 4.83 for SciLate, and 4.96 for SciFresh (corresponding to relative errors of approximately 8.02 %, 7.56 %, 6.78 %, and 6.27 %, respectively). These values indicate that YOLO11n demonstrates high precision in fruitlet counting, with an error margin below 8 % across all varieties. The MAE results further illustrate its precision, with the lowest values recorded at 4.07 for Honeycrisp, 3.98 for Cosmic Crisp, and 3.85 for SciFresh. The exception was SciLate, where YOLO11n recorded a higher MAE of 7.73, suggesting a potential area for improvement. Likewise, in images captured by the Intel RealSense machine-vision camera, YOLO11n consistently demonstrated the most accurate performance. For SciLate, it achieved the lowest RMSE of 4.83 (≈6.78 % error), but a comparatively higher MAE of 7.73 (≈10.86 % error). This combination reflects the model's precise yet slightly inconsistent counting ability under certain conditions. In contrast, for the SciFresh variety, YOLO11n again recorded the lowest RMSE of 4.96 (≈6.27 % error) and an MAE of 3.85 (≈4.87 % error), showcasing its robustness in consistently and accurately detecting fruitlets. These results highlight YOLO11n's ability to generalize across sensor types and maintain strong performance, particularly in real-time detection and counting tasks under variable orchard conditions.

(4) Sensor-Specific Training Impact. The study demonstrated that models trained exclusively on data from a specific sensor, such as the Intel RealSense, exhibited higher performance when evaluated on data from the same sensor compared to data from other sensors. This finding highlights the critical importance of incorporating diverse sensor data during training to achieve robust model performance across varied deployment scenarios.
(5) Inference Speed for Deployment. When deploying YOLO models in automation and robotics, particularly for real-time agricultural tasks such as early-stage crop load management in apple orchards, the choice of model configuration becomes crucial. Each YOLO family includes configurations optimized for differing balances of speed and accuracy. For example, YOLOv8n was the fastest in its family with an inference speed of 4.1 ms, making it suitable for applications with limited computational resources. In the YOLOv9 series, YOLOv9 GELAN-s achieved the fastest speed at 11.5 ms. Among YOLOv10 configurations, YOLOv10n achieved the best inference speed of 5.5 ms, while YOLOv12n led its family with a speed of 4.6. The fastest model configuration across all tested versions was YOLO11n, with an inference speed of 2.4 ms, making it the optimal choice for environments where both high speed and precision are critical for efficient automation and real-time decision-making.

This analysis leaves room for future exploration of emerging models like YOLOv13 and YOLO26 (or YOLOv26) to further benchmark and compare advancements in real-time detection accuracy and efficiency within YOLO framework.

## CRediT authorship contribution statement

**Ranjan Sapkota:** Writing – review & editing, Writing – original draft, Visualization, Validation, Software, Methodology, Investigation, Formal analysis, Data curation, Conceptualization. **Zhichao Meng:** Writing – review & editing, Software, Methodology. **Martin Churuvija:** Writing – review & editing, Methodology. **Xiaoqiang Du:** Writing – review & editing, Methodology. **Zenghong Ma:** Software, Conceptualization. **Manoj Karkee:** Writing – review & editing, Writing – original draft, Visualization, Supervision, Resources, Project administration, Funding acquisition.

## Data availability

The data and source code related to this study are available from the corresponding author upon reasonable request.

## Declaration of competing interest

The authors declare that they have no known competing financial interests or personal relationships that could have appeared to influence the work reported in this paper.

## Acknowledgement

This work was supported in part by the National Science Foundation (NSF) and the United States Department of Agriculture (USDA), National Institute of Food and Agriculture (NIFA), through the “Artificial Intelligence (AI) Institute for Agriculture” program under Award Numbers AWD003473 and AWD004595, and USDA-NIFA Accession Number 1029004 for the project titled “Robotic Blossom Thinning with Soft Manipulators.” Additional support was provided through USDA-NIFA Grant Number 2024-67022-41788, Accession Number 1031712, under the project “ExPanding UCF AI Research To Novel Agricultural EngineeRing Applications (PARTNER)”. We acknowledge Allan Brothers Fruit Company for granting access to their commercial orchards across Washington State, USA. Special thanks to Astrid Wimmer, Randall Cason, Diego Lopez, Giulio Diracca, and Priyanka Upadhyaya for their invaluable efforts in data preparation and logistical support throughout this project.